\documentclass[letterpaper, 10 pt, conference]{ieeeconf}  

\IEEEoverridecommandlockouts                              

\usepackage{url}
\usepackage{tikz}
\usepackage{subcaption}

\title{\LARGE \bf
XBG: End-to-end Imitation Learning for Autonomous Behaviour in Human-Robot Interaction and Collaboration
}

\author{Carlos Cardenas-Perez$^{1}$, Giulio Romualdi$^{1}$, Mohamed Elobaid$^{1}$, Stefano Dafarra$^{1}$, Giuseppe L'Erario$^{1}$, \\ Silvio Traversaro$^{1}$, Pietro Morerio$^{2}$, Alessio {Del Bue}$^{2}$ and Daniele Pucci$^{1,3}$
\thanks{*The paper was supported by the Italian National Institute for Insurance
against Accidents at Work (INAIL) ergoCub Project.}
\thanks{$^{1}$Artificial and Mechanical Intelligence, Istituto Italiano di Tecnologia, Genoa, Italy.
        }
\thanks{$^{2}$Pattern Analysis and Computer Vision, Istituto Italiano di Tecnologia, Genoa, Italy.
        }
\thanks{$^{3}$ Department of Computer Science, The University of Manchester, Manchester, United Kingdom}
}

\begin{document}

\maketitle
\thispagestyle{empty}
\pagestyle{empty}

\begin{abstract}
This paper presents XBG (eXteroceptive Behaviour Generation), a multimodal end-to-end Imitation Learning (IL) system for a whole-body autonomous humanoid robot used in real-world Human-Robot Interaction (HRI) scenarios.
The main contribution of this paper is an architecture for learning HRI behaviours using a data-driven approach. 
Through teleoperation, a diverse dataset is collected, comprising demonstrations across multiple HRI scenarios, including handshaking, handwaving, payload reception, walking, and walking with a payload. 
After synchronizing, filtering, and transforming the data, different Deep Neural Networks (DNN) models are trained. 
The final system integrates different modalities comprising exteroceptive and proprioceptive sources of information to provide the robot with an understanding of its environment and its own actions. 
The robot takes sequence of images (RGB and depth) and joints state information during the interactions and then reacts accordingly, demonstrating learned behaviours. By fusing multimodal signals in time, we encode new autonomous capabilities into the robotic platform, allowing the understanding of context changes over time. 
The models are deployed on ergoCub, a real-world humanoid robot, and their performance is measured by calculating the success rate of the robot's behaviour under the mentioned scenarios.
\end{abstract}
\section{INTRODUCTION}
Humanoids robots possess a structure that offers operational flexibility enabling them to excel in tasks demanding intricate mobility and manipulation skills that aligns closely with environments intended for humans \cite{Darvish2023teleoperation}. Thanks to advances in both hardware and software, robots are moving from standalone machines to collaborative systems \cite{SEMERARO2023102432}, having a positive impact in many fields such as manufacturing, nursing, social, among others \cite{abaigbena2024aihri}. This paper contributes towards the development of autonomous humanoid robots combining data driven approaches with whole-body control for Human-Robot Interaction (HRI) tasks.

For an effective Human-Robot Interaction and Collaboration (HRIC), the robot must engage in a safe physical interaction both with the user and the environment elements, comprehend the human's intentions, and determine when to take the initiative or assist the human in different situations \cite{rozo2016controllers}. Achieving such collaboration autonomously poses a significant challenge, leading many works to focus on teleoperation strategies, integrating human cognitive capabilities (operator intelligence) with the robot's physical abilities \cite{luo2023104414}. 

Teleoperation systems, beyond enabling humans to control robots and leverage their physical capabilities, also allow recording demonstrations by human experts, opening the possibility to learn control and behaviours policies from such examples. Many works in the literature refers to this demonstration learning as Imitation Learning (IL). IL schemes usually consists on learning to predict the robot motor states in a future time step based on the current or past sensory-motor state \cite{ogata2022cognitive}. Learning policies from demonstrations capable to map from images to actions have shown good results in different fields of robotics \cite{Zhang2017DeepIL}, like autonomous driving \cite{codevilla2022mutimodal,perez2022visually} and serial manipulators policies \cite{sergey2018vision, Mandlekar2020git, nasiriany2022learning, finn2017oneshot}. 

\begin{figure}
    \includegraphics[width =0.5\textwidth]{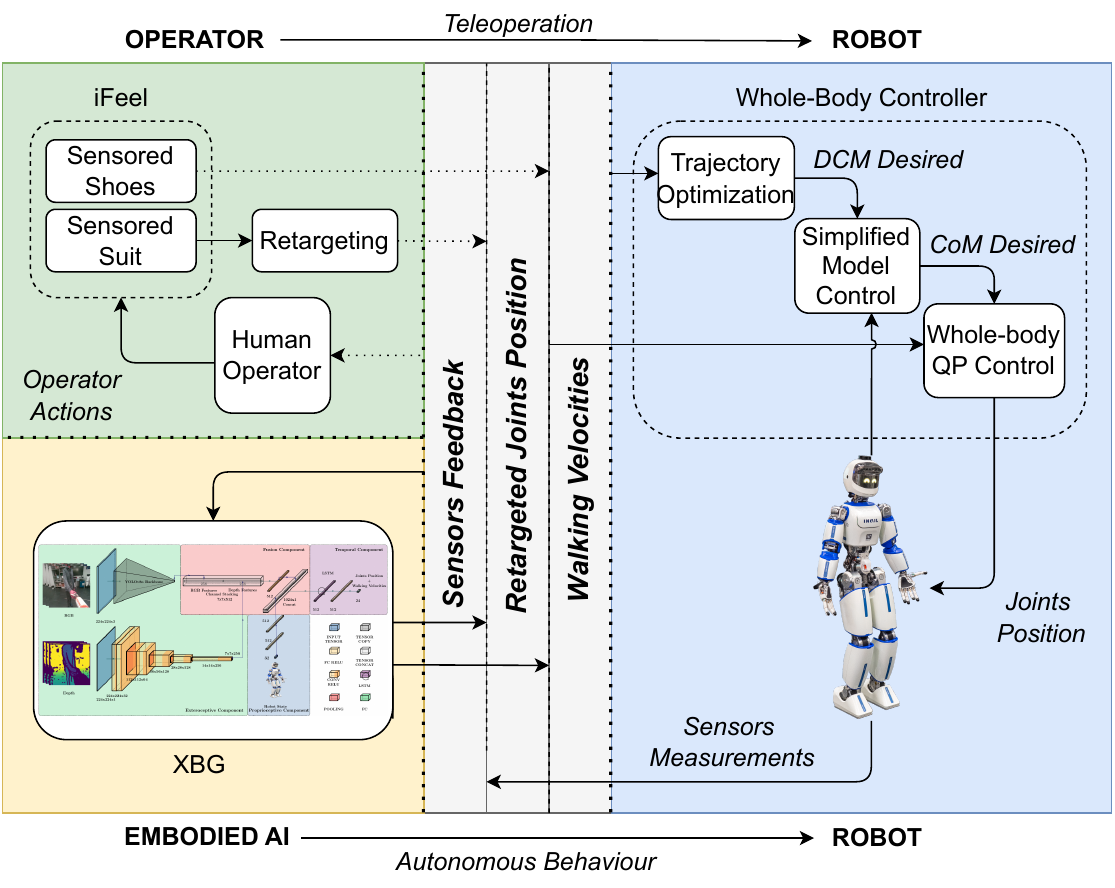}
     \caption{From Teleoperation to Autonomy: the robot architecture uses the retargeted joints positions and walking velocities interfaces and provides the information for the sensors feedback interface. The robot can be operated either from a human operator or from XBG.}
    \label{fig:xbgSystem}
\end{figure}

Different learning from demonstration strategies, called Behaviour Cloning (BC), were compared in   \cite{mandlekar2021whatmatters}, in a set of manipulation tasks for a robotic arm. The study concludes that models considering temporal dependencies present a better performance than those which only use a single time step state. 
Inspired by ViT \cite{Dosovitskiy2020AnII}, some works start to include transformer architectures in their IL systems. VIOLA \cite{zhu2022viola} aims for an object centric IL policy that uses a transformer where its feature encoding is composed by concatenating feature maps extracted from both workspace image and eye-in-hand image, K region features (which are bounding boxes positions from a parallel object detection branch conducted in the workspace image) and propioceptive features obtained from the robot state. VIOLA obtains an IL policy for six different manipulation tasks.

Multimodal end-to-end systems for IL exist in the state of the art. For instance, one notable study presents an autonomous driving system that utilizes RGB, depth, and vehicle measurements as inputs \cite{codevilla2022mutimodal}. This system can map these modalities directly to steering angle, throttle, and brake actions. Various modality fusion strategies are proposed and tested using the CARLA simulator, which is also the platform used to collect the training data. A high-level planner issues navigation commands by selecting different branches in the network corresponding to each command available in the training data.

Recent works have started to explore learning humanoid robot manipulation based on demonstrations \cite{seo2023trill}. A visuomotor policy for a humanoid robot is trained using a Virtual Reality (VR) headset and VR hand controllers. The system outputs the cartesian set points for the hand frames, a grasping signal and predefined gait commands. Nonetheless, this work focus on manipulation tasks and not on HRIC, additionally there is no locomotion presented on the real robot. 

Regarding human-robot interaction and collaboration, a review study \cite{robinson2023hric} reveals that only 3\% of the HRIC related papers are leveraging learning from demonstrations. Most of the existing works focus on gesture and action recognition, usually aiming for states identification and most of them are either mobile platforms or fixed manipulators. Authors in \cite{seo2023trill} also point out the complexity of collecting demonstrations for humanoid robots, specifically because of their floating-base dynamics, involving stability and state estimation challenges. Additionally, in HRIC scenarios the robot needs to learn to generate its actions but also understand how others actions affect itself, adapting its behaviour to those effects \cite{ogata2022cognitive}. 

This article presents a behaviour generation system for humanoid robots, based on imitation learning, moving from teleoperation to autonomy. The main contributions of this work are:
\begin{itemize}
    \item XBG (eXteroceptive Behaviour Generation): An Imitation Learning approach that maps perception to robot motion for humanoid behaviour generation.
    \item A training pipeline that uses teleoperation data to develop autonomous behaviours, with a focus on Human-Robot Interaction and Collaboration (HRIC).
    \item Validation and ablation tests of XBG conducted on the real humanoid robot platform, ergoCub.
\end{itemize}

This paper is organized as follows: section \ref{BACKGROUND} presents the existing teleoperation system, section \ref{XBGSYSTEM} presents the architecture of the network and explains the intuition of its components, section \ref{EXPERIMENTS} provides the details of the experiments conducted and the results obtained on this work, finally section \ref{CONCLUSIONS} concludes the paper and proposes some future works.

\section{BACKGROUND}\label{BACKGROUND}
XBG leverages the avatar system \cite{dafarra2024icub3} that enables a human operator to embody a humanoid robotic platform. It is composed of two architectures: the operator and the robot one, which are connected through diverse teleoperation and teleperception interfaces. Figure \ref{fig:xbgSystem} shows the operator (left-top) and the robot architecture (right) connected through the retargeted joints position, walking velocities, and the sensor feedback interface. We introduce the embodied AI architecture (left-bottom) where XBG model connects to the same interfaces as the operator one allowing to switch from human teleoperation to an embodied AI autonomous behaviour (the XBG neural network architecture is presented in detail in figure \ref{fig:architecture}). 

\subsection{Operator Architecture}
 The operator architecture consists of three main components, the human operator, the motion capture system, and the whole-body retargeting as shown in figure \ref{fig:xbgSystem} (left-top). The human operator receives feedback from the robot sensors (visual, auditory, and haptic) allowing they to understand the robot environment from a first person point of view and then provide the decision making. 
 
 The second measures and registers the operator's movements using a set of wearable devices and Virtual Reality (VR) equipments. The operator wears the iFeel \cite{iFeel} equipment, a gear composed of a sensorized suit and shoes that allow us to capture his movements and then retarget them to the robot using the Whole-body geometric retargeting \cite{kourosh2019retargeting}.
 
 This retargeting module uses inverse kinematic methods to map human joint angles and velocities to the humanoid robot joints. It only requires the human links' rotation and angular velocities, and robot's URDF model (Unified Robot Description Format). Even when the human joints configuration are not identical to the robot ones, the retargeting conducts a dynamical optimization that aims to minimize the rotation error between the human and robot frames, considering the robot balance, manoeuvrability and the robot limitations.  
 
\subsection{Robot Architecture}

 The robot architecture presents a whole-body controller for locomotion of bipedal humanoid robots. The robot control uses a layered control architecture \cite{romualdi2020benchmarking}, which is composed of three layers: trajectory optimization, simplified model control, and whole-body quadratic programming (QP) control. Each layer uses signals from the robot state, the output of the previous layer and/or reference signals from the operator. Initially the trajectory optimization works as a footstep planner calculating the desired feet trajectory, contact points and timing. Then, the simplified model control looks for a feasible Center of Mass (CoM) trajectory which is received by the whole-body controller together with the retargeted joint position as a regularization and generates the desired positions, velocities or torques for the robot joints.

\begin{figure*}[ht]
    \includegraphics[width =\textwidth]{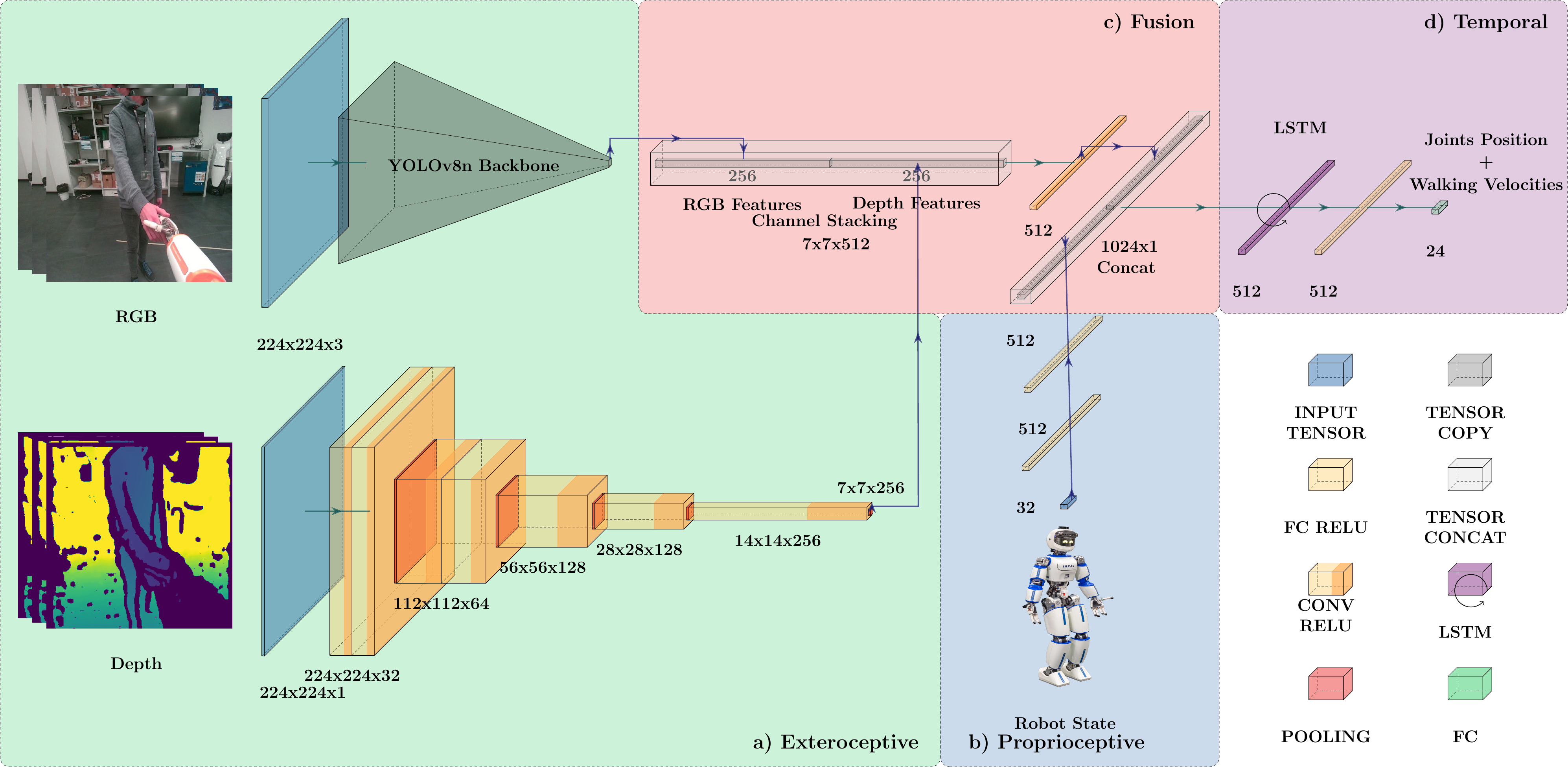}
     \caption{XBG Neural Network Architecture. a) Exteroceptive Component comprising RGB branch (top) and Depth branch (bottom). b) Propioceptive Component. c) Modal Fusion Component. d) Temporal Component.}
    \label{fig:architecture}
\end{figure*}

\section{XBG Model}\label{XBGSYSTEM}

In an autonomous humanoid robot system, much like humans being aware of their surroundings and actions, it is essential for the robot to understand its environment and its own actions. To achieve this, we incorporate two components into the XBG (eXteroceptive Behaviour Generation) model one for processing environmental information (such as what the robot sees and senses) and another for understanding the robot's state (like its position, movements, and internal conditions). Similar to humans, who do not make decisions or act based on a single moment in time but rather consider longer context to give a sense to an interaction, the robot also needs to understand sequences of events. XBG analyzes the images captured by the robot's cameras and monitor its joint positions over time to assess its behaviour. This involves observing changes over time, both in visual perception and physical movements.

To this purpose the model comprises 4 primary components that are shown in Figure \ref{fig:architecture}: one for extracting features from the environmental sensory inputs (exteroceptive), a component to assess its state (proprioceptive), another one for integrating these diverse sources of information (modality fusion), and the last one for analyzing temporal dependencies and patterns. These components collaborate to provide the robot a comprehensive understanding of how it should interact with its surroundings and humans. 

\subsection{Exteroceptive Component}
The exteroceptive component aims at enhancing the robot's perception capabilities. By gathering information from the surrounding environment, the robot gains insight into its context and learns to respond accordingly. For such a task, this work uses both RGB and depth data as shown in Figure \ref{fig:architecture}a. Each modality is initially handled by separate branches, extracting a set of features from them, that are then fused.  

The RGB branch receives input images of size 224x224 pixels by 3 channels. These images go through a feature extraction using a pre-trained model backbone. XBG uses the smallest version of YOLOv8 (You Only Look Once Version 8 nano) \cite{yolov8_ultralytics}. The purpose of using this pretrained sub-network is to leverage its capacity to extract a meaningful set of visual features. 

On the other hand, the depth branch employs a series of 7 standard convolutional layers, each layer uses a 3x3 kernel, with stride and padding set to 1, and ReLU activation function. The number of channels increase gradually up to 256 throughout the branch, while its dimensionality is reduced by using pooling layers. This branch's design ensures an output with identical dimensions to the RGB branch, facilitating the multimodal fusion in subsequent network layers.

\subsection{Proprioceptive Component} 
The proprioceptive component plays a crucial role in furnishing the robot with self-awareness by extracting pertinent information regarding its own actions. It encompasses the robot's upper-body joint positions, locomotion velocities (including linear, angular, and sideways velocities), and motor currents in the arm joints (shoulders and elbows), which provide insight into the effort exerted by these joints. 
Figure \ref{fig:joints} shows the degrees of freedom of each joint controlled by XBG in the upper-body of the robot and the walking velocities signal for the walking controller in charge of the lower-body joints.
These inputs are fed into two fully-connected layers, each comprising 512 neurons with ReLU activation function, obtaining a robot state feature tensor.

\subsection{Modal Fusion Component}
XBG employs a mid-fusion strategy to fuse the various modalities within the system (figure \ref{fig:architecture}c), it starts by stacking the channels from the RGB and depth branch conforming a 7x7x512 tensor which pass through a convolutional layer with kernel size of 7. This allows to explicitly provide some spatial relationship between the RGB and the depth information and also reduce the dimensionality of the exteroceptive component obtaining a 512x1x1 already flattened tensor.

This tensor contains the exteroceptive features that are concatenated with the proprioceptive features, obtaining a 1024 length tensor that constitutes the whole spatial features for a single instant of time.

\subsection{Temporal Component}
Thanks to the exteroceptive and proprioceptive components, the robot possesses a set of spatially related features that map the external observation of the environment and the internal robot state. To construct a successful autonomous system, it's crucial to account for both spatial and temporal dimensions, specifically understanding how the environment evolves over time and how the robot reacts accordingly. To capture these temporal relationships, XBG utilizes a recursive neural network layer, specifically a Long Short Term Memory (LSTM) with a hidden size of 512 and considers a time window of 1.6 seconds during training. Finally, the network includes a non-linear fully connected layer followed by a linear one, forming the network's head. This allows us to decode the features extracted directly to the joint positions for the entire upper body and the locomotion velocities for the walking controller in charge of the lower-body of the robot.

\section{EXPERIMENTS}\label{EXPERIMENTS}
\subsection{Experimental Setup}
We used an extension of the avatar system described in the background section enabling full embodiment and manipulation of ergoCub robot \cite{ergoCub}, the humanoid robotic platform used for the experiments. Additionally to the iFeel suit we use the SenseGlove DKI haptic gloves to control the robot hands and the HTC VIVE Pro Eye headset which provides the orientation of the operator head and gaze tilt, it also allows the operator to receive the visual feedback from the robot's cameras (Realsense D450) placed at the robot's head providing the robot point of view image.

\begin{figure}[ht]
    \includegraphics[width =0.45\textwidth]{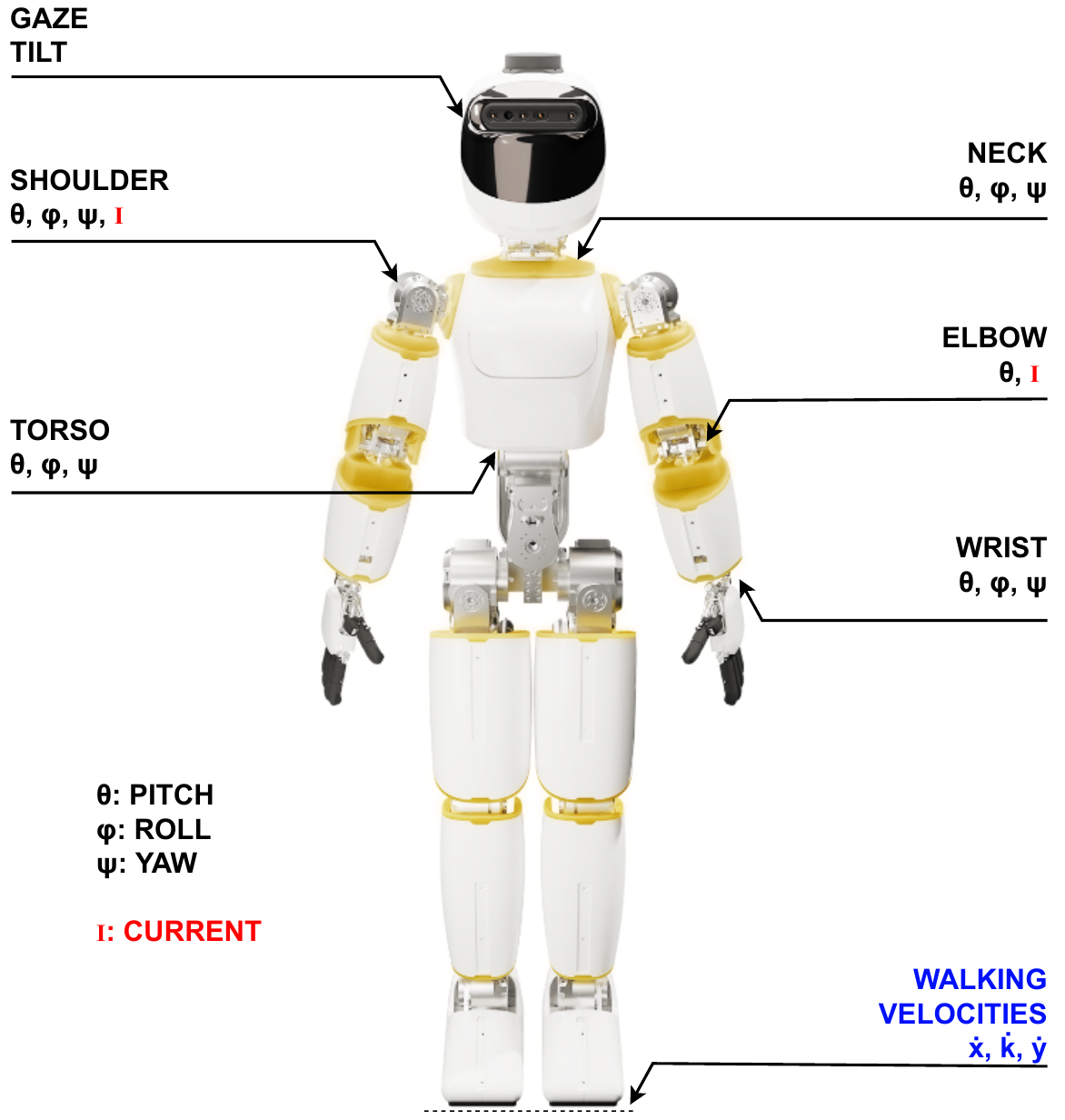}
     \caption{XBG proprioceptive input/output signals. XBG receives as input the positions from the upper-body joints, motor currents from shoulders and elbows, and walking velocities. It outputs the same joints positions for the upper-body and the walking velocities for the walking controller.}
    \label{fig:joints}
\end{figure}

\subsection{Data Collection}

The dataset compiled for this study comprises 100 minutes of the robot's behaviour demonstrations while being teleoperated in various Human-Robot Interaction (HRI) scenarios:

\subsubsection{Handwave} The human raises both arms and wave them to greet the robot. The robot should raise the arms to greet back (In some examples only one arm was raised during the interaction).
\subsubsection{Handshake} The human approaches the robot from different angles and extends his hand to shake hands with the robot. The robot should extends one of its arms until touching the human hand to shake hands.
\subsubsection{Payload reception} The human approaches the robot from different angles and extends a payload to the robot. The robot should extends his arms to receive the payload and then hold it until the human takes it back.
\subsubsection{Walk / Person following} The human walks away from the robot moving both arms towards his body like calling the robot, signaling the robot to walk and follow the person.
\subsubsection{Walk with Payload}  Once the human delivers a payload to the robot, he can signal the robot to walk while carrying the payload using a similar movement as the one describe in the "Walk" behaviour. The robot should be able to walk balancing the payload and not letting it fall.

\begin{figure*}
\begin{minipage}[t]{\textwidth}
\begin{subfigure}[t]{0.32\linewidth}%
    \includegraphics[width=\linewidth]{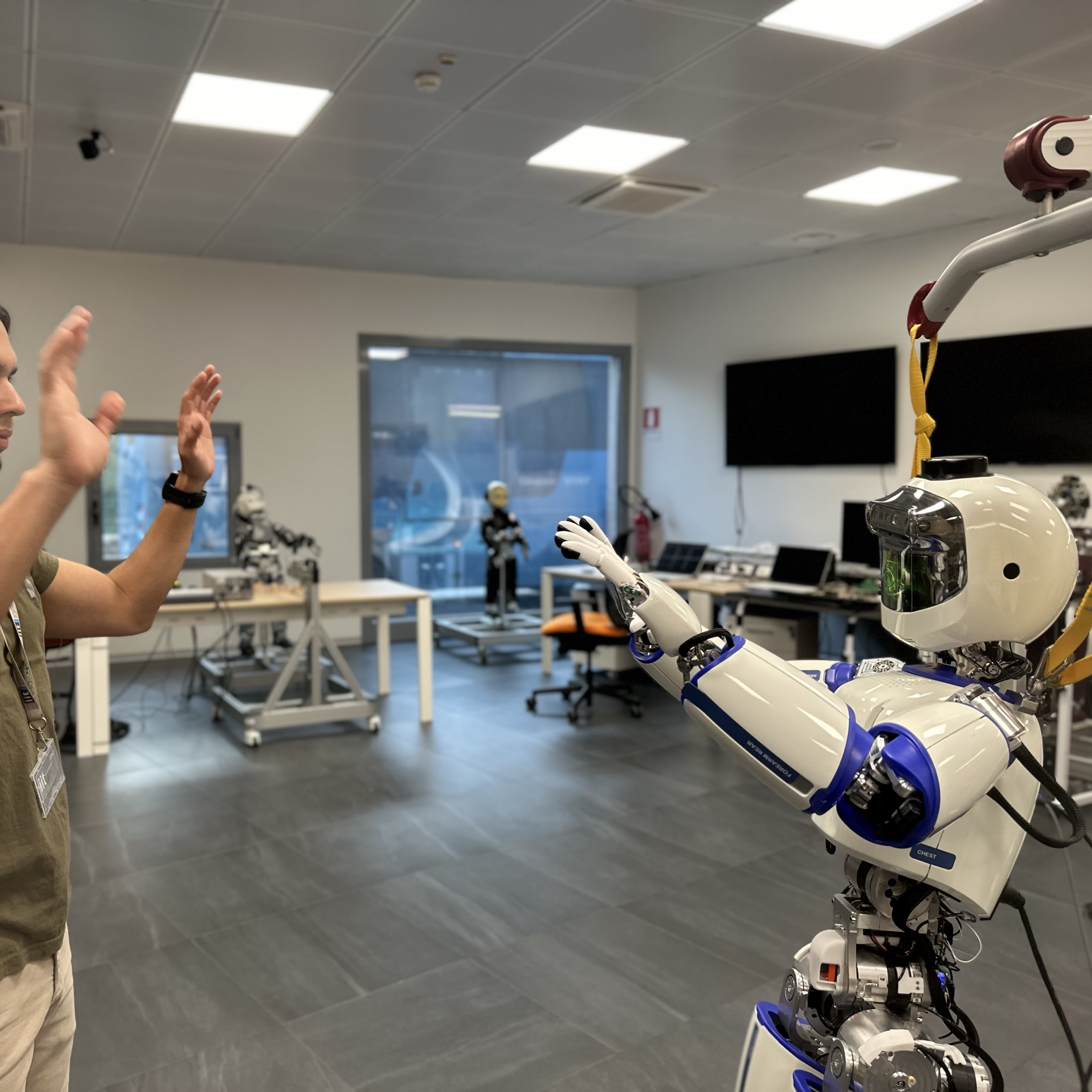}
    \caption{Handwaving}
\end{subfigure}
\begin{subfigure}[t]{0.32\linewidth}%
    \includegraphics[width=\linewidth]{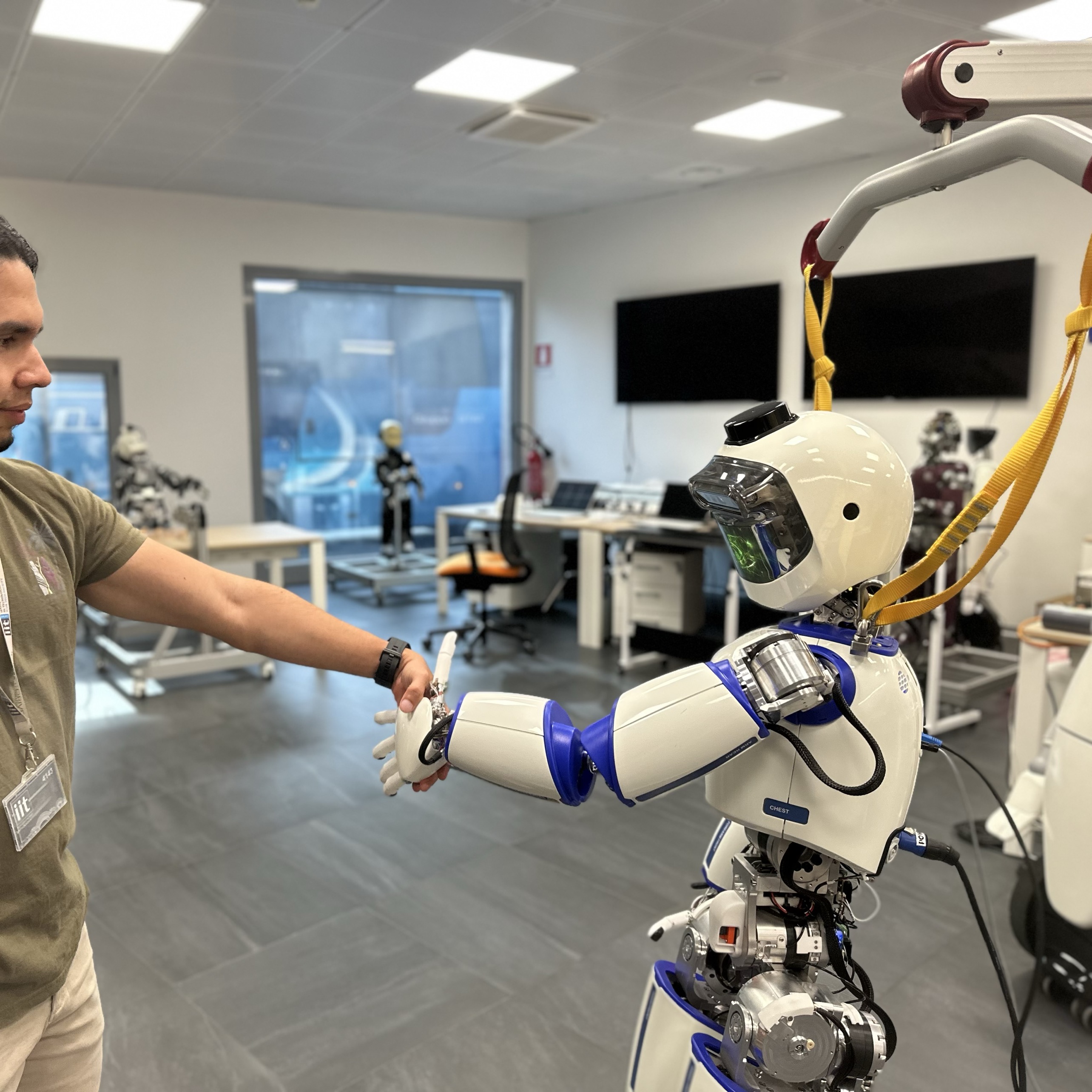}
    \caption{Handshaking}
\end{subfigure}
\begin{subfigure}[t]{0.32\linewidth}%
    \includegraphics[width=\linewidth]{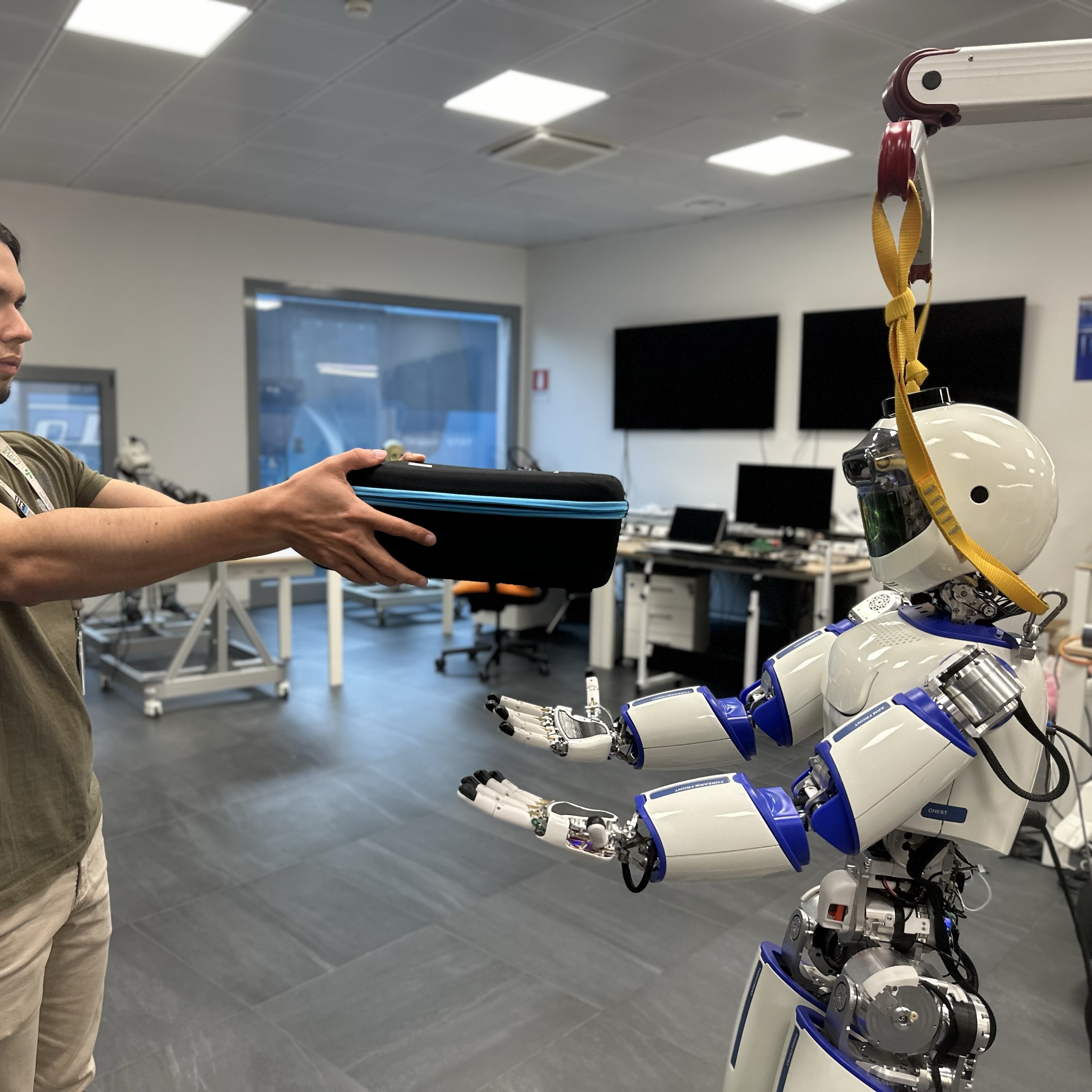}
    \caption{Payload reception}
\end{subfigure}
\begin{subfigure}[t]{0.66\linewidth}%
    \includegraphics[width=0.485\linewidth]{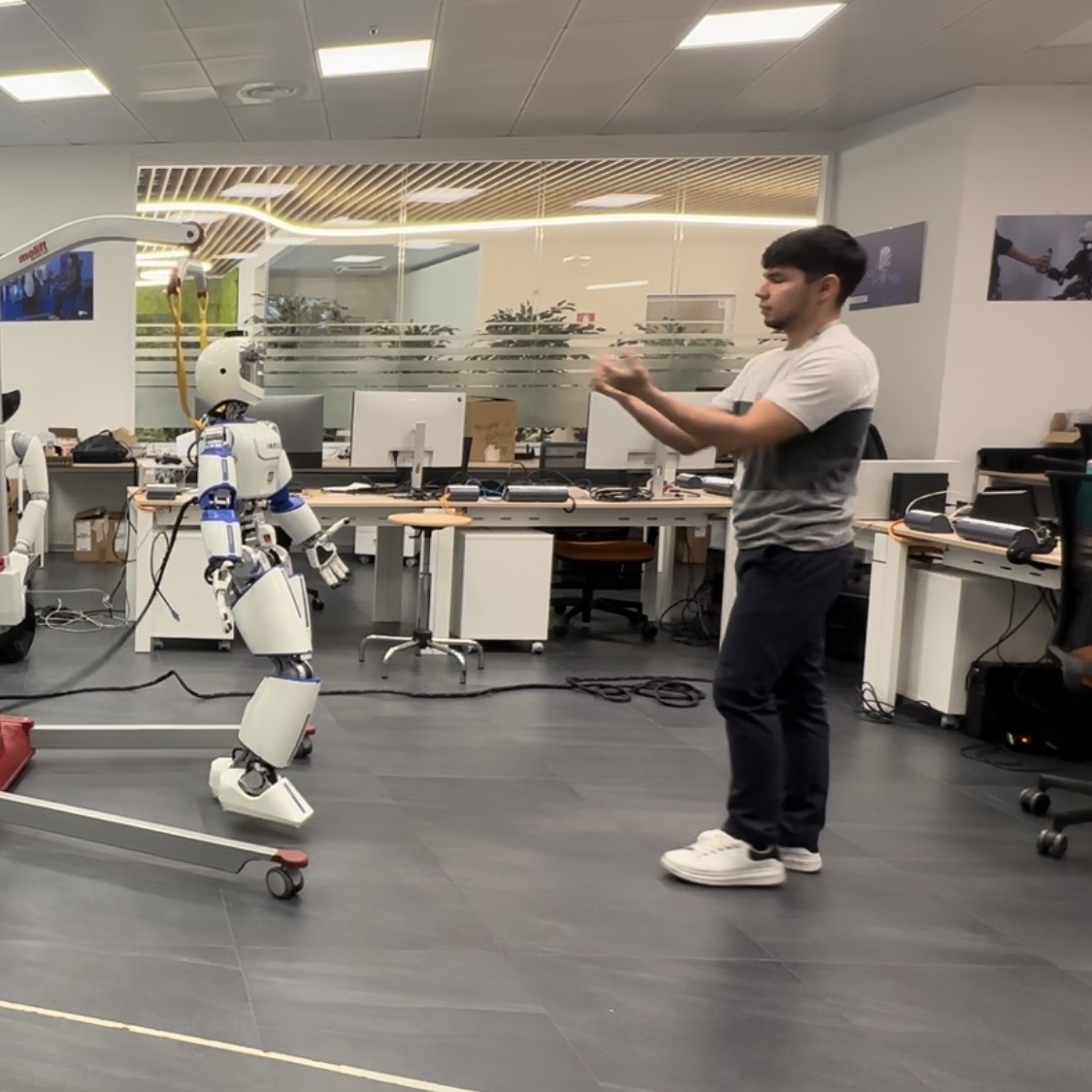}%
    \hfill
    \includegraphics[width=0.485\linewidth]{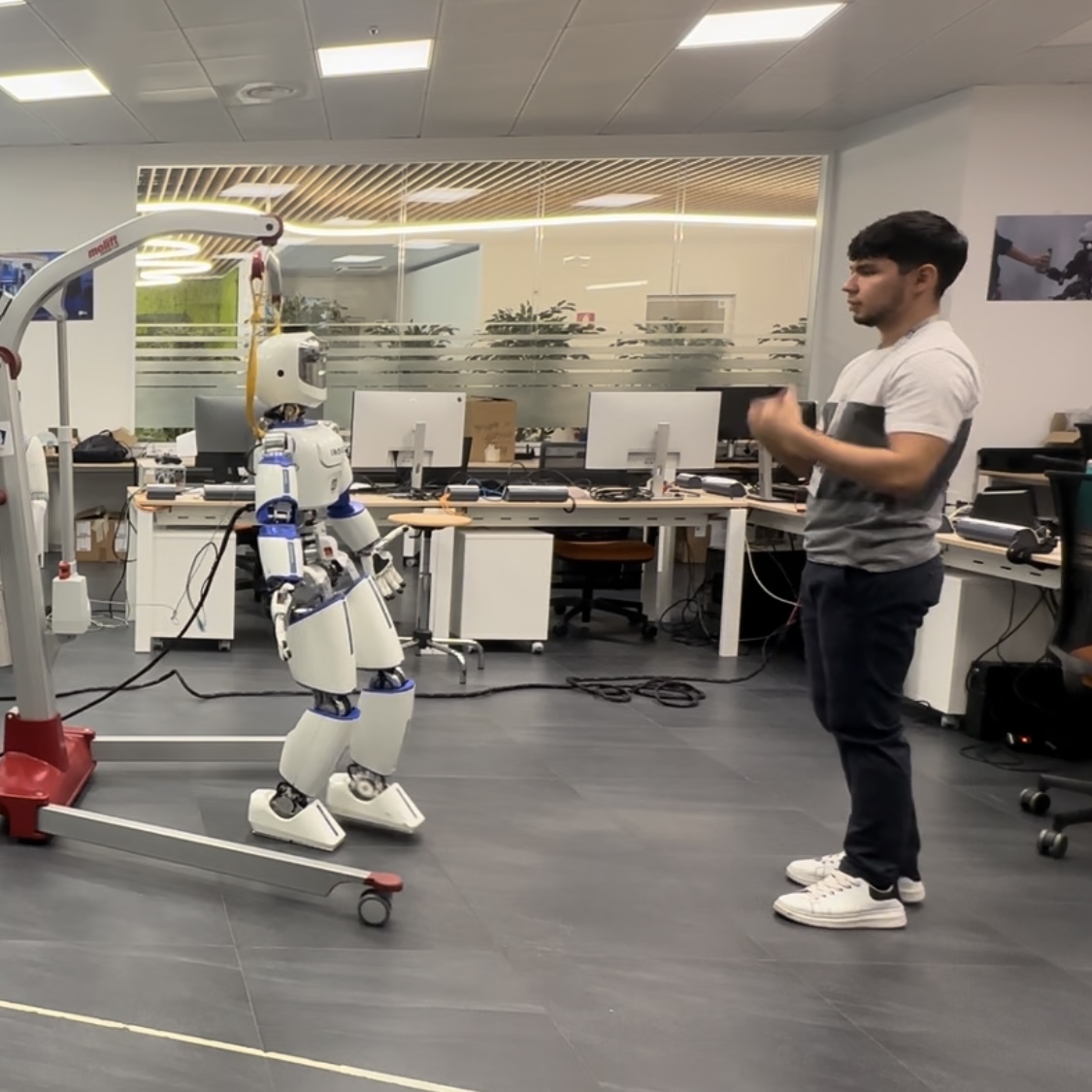}
    \caption{Walking}
\end{subfigure}\hfill
\begin{subfigure}[t]{0.32\linewidth}%
    \includegraphics[width=\linewidth]{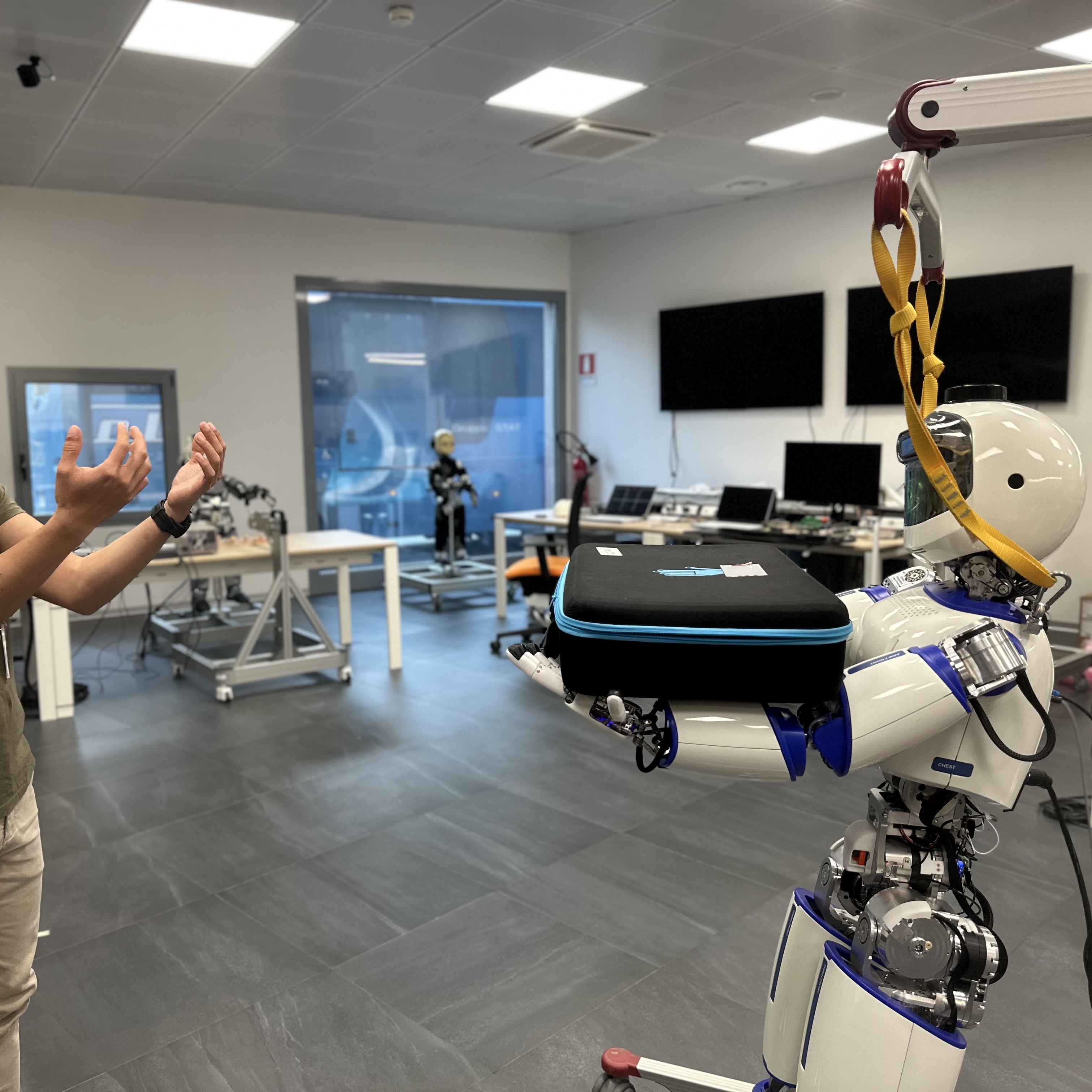}
    \caption{Walking with payload}
\end{subfigure}

\end{minipage}
\caption{Human-Robot Interaction scenarios presented in the XBG system.}\label{fig:behaviours}
\end{figure*}

Figure \ref{fig:behaviours} provides an example of the different interaction scenarios considered for this work. The dataset encompasses examples involving three different operators and interactions with approximately ten different individuals. Notably, during both data collection and testing, only one person interacted with the robot at a time, although there were instances where multiple people were within the robot's field of view.
The logging system captures depth and RGB images from the robot's cameras at a frequency of 30Hz. Simultaneously, another thread logs the robot's state (joint positions, velocities, accelerations, motor temperatures and currents, among other sensors values) at a frequency of 100Hz while this interaction between the human and the robot occurs.

\subsection{Data Processing}
To construct the dataset, we need to create training sequences and synchronize the robot's joint state with the camera data. However, we noticed that consecutive images at 30Hz often exhibit minimal variation, resulting in lengthy sequences with redundant information. Consequently, we down-sample the data to 10Hz and establish a 1.6-second time window, reducing the number of samples while maintaining a balance between providing temporal context to the network and meeting computational requirements for both training and online deployment on the robot. This down-sampling enhances the significance of information between consecutive time steps within a sample sequence and facilitates synchronization with the sensors data. 

In addition to the signals presented in Figure \ref{fig:joints}, we also incorporate the electrical current readings from the arm motors (shoulders and elbows) as they can aid the neural network in understanding the effort exerted by these robot's joints.
However, these signals often contain a high level of noise, leading to shaky movements in the robot's behaviour. To address this, we apply a second-order Butterworth low-pass filter to the arm motor currents as part of the preprocessing before getting into the network, reducing noise and resulting in smoother movements in the robot. Regarding the depth information we decided to encode the distance between 0 and 1 values, since the robot is intended to work in human-robot interaction scenarios, we clamp the distance with a maximum of 4 meters, corresponding to the value 1.

\begin{figure*}[ht]
    \begin{subfigure}[t]{0.121\linewidth}%
        \includegraphics[width=\textwidth]{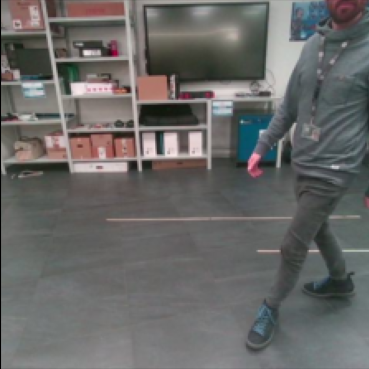}
    \caption{t=0.1 s}\end{subfigure}
    \begin{subfigure}[t]{0.121\linewidth}
        \includegraphics[width=\textwidth]{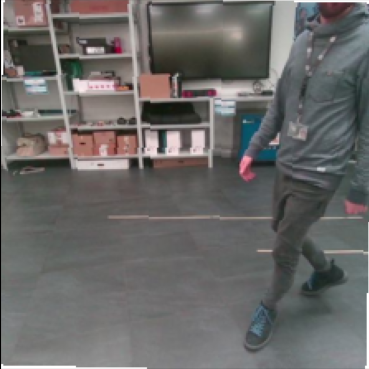}
    \caption{t=0.2 s}\end{subfigure}
    \begin{subfigure}[t]{0.121\linewidth}
        \includegraphics[width=\textwidth]{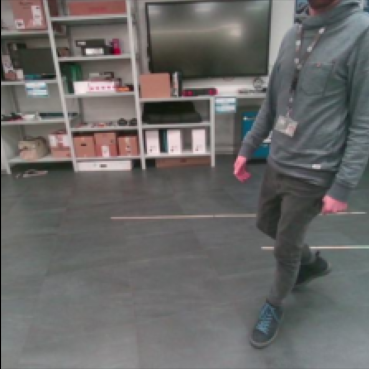}
    \caption{t=0.3 s}\end{subfigure}
    \begin{subfigure}[t]{0.121\linewidth}
        \includegraphics[width=\textwidth]{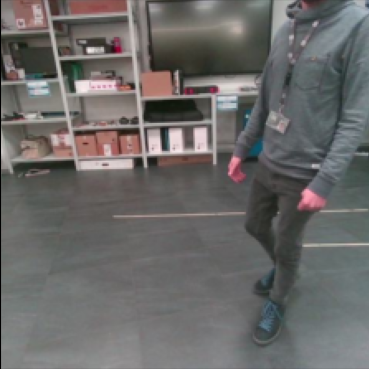}
    \caption{t=0.4 s}\end{subfigure}
    \begin{subfigure}[t]{0.121\linewidth}
        \includegraphics[width=\textwidth]{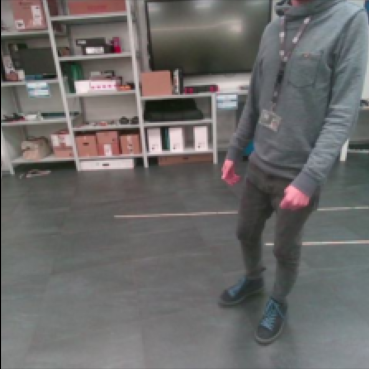}
    \caption{t=0.5 s}\end{subfigure}
    \begin{subfigure}[t]{0.121\linewidth}
        \includegraphics[width=\textwidth]{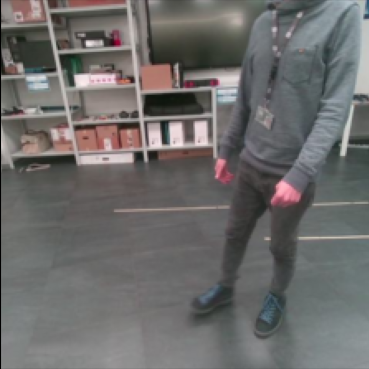}
    \caption{t=0.6 s}\end{subfigure}
    \begin{subfigure}[t]{0.121\linewidth}
        \includegraphics[width=\textwidth]{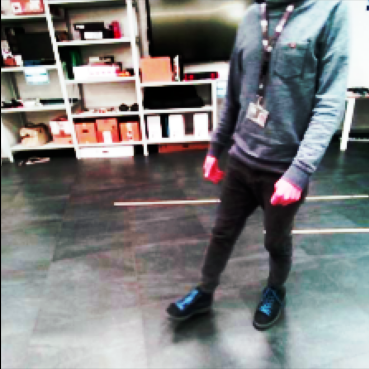}
    \caption{t=0.7 s}\end{subfigure}
    \vspace{2pt}
    \begin{subfigure}[t]{0.121\linewidth}
        \includegraphics[width=\textwidth]{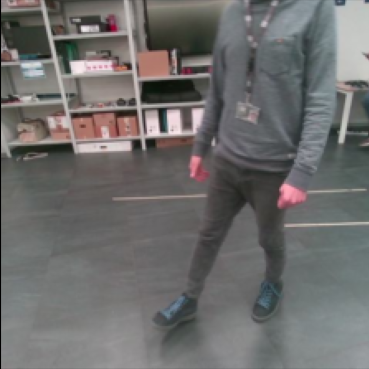}
    \caption{t=0.8 s}\end{subfigure}
    \begin{subfigure}[t]{0.121\linewidth}
        \includegraphics[width=\textwidth]{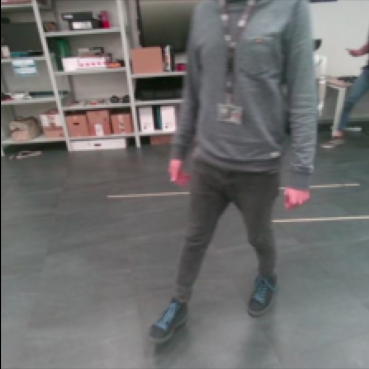}
    \caption{t=0.9 s}\end{subfigure}
    \begin{subfigure}[t]{0.121\linewidth}
        \includegraphics[width=\textwidth]{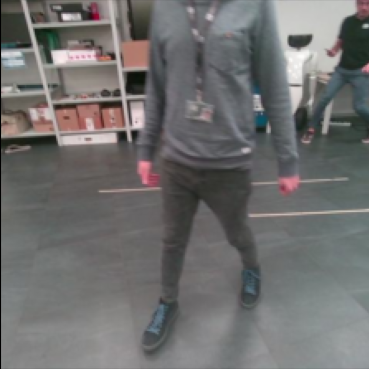}
    \caption{t=1.0 s}\end{subfigure}
    \begin{subfigure}[t]{0.121\linewidth}
        \includegraphics[width=\textwidth]{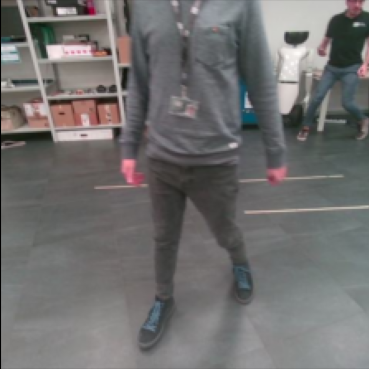}
    \caption{t=1.1 s}\end{subfigure}
    \begin{subfigure}[t]{0.121\linewidth}
        \includegraphics[width=\textwidth]{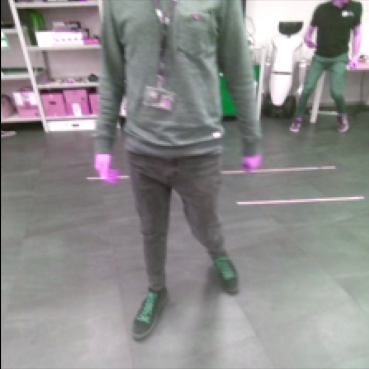}
    \caption{t=1.2 s}\end{subfigure}
    \begin{subfigure}[t]{0.121\linewidth}
        \includegraphics[width=\textwidth]{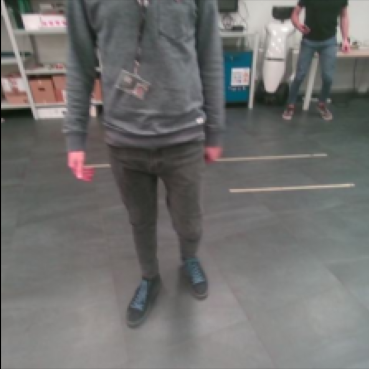}
    \caption{t=1.3 s}\end{subfigure}
    \begin{subfigure}[t]{0.121\linewidth}
        \includegraphics[width=\textwidth]{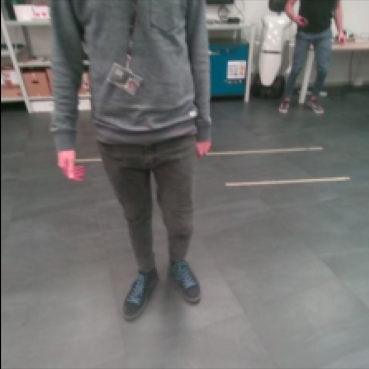}
    \caption{t=1.4 s}\end{subfigure}
    \begin{subfigure}[t]{0.121\linewidth}
        \includegraphics[width=\textwidth]{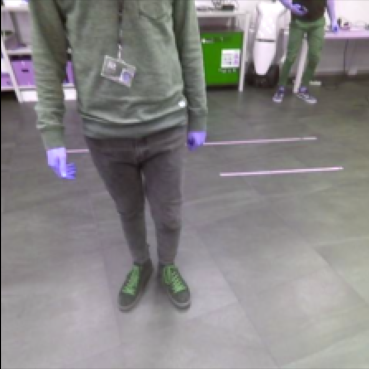}
    \caption{t=1.5 s}\end{subfigure}
    \begin{subfigure}[t]{0.121\linewidth}
        \includegraphics[width=\textwidth]{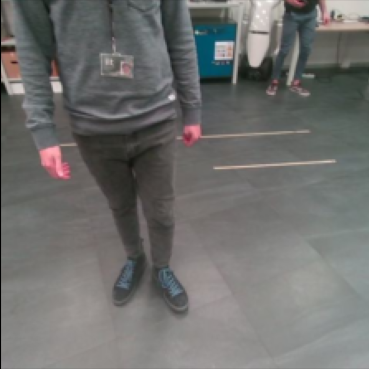}
    \caption{t=1.6 s}\end{subfigure}
    \caption{RGB Sequence sample for training. 1.6 seconds time window sequence at 10Hz with augmentations used for training the network.}\label{fig:sequence}
\end{figure*}

Lastly, in an effort to enhance the network's generalization ability, we apply various augmentations to the RGB images during training time. These include random erasing, random zoom out, elastic transform, random equalizer, color jitter, gaussian blur, and random auto-contrast. Each frame within a sequence has a 0.3 probability of undergoing one of these transformations. This approach helps prevent the network from overfitting to specific lighting conditions, facial features, clothing color, and makes it more robust to potential noise in the images. Figure \ref{fig:sequence} illustrates an example of a sequence of RGB data seen by the network after the down-sampling, synchronization, and augmentations. Following the IL paradigm, the output of the network are the joint positions for the entire upper body and the locomotion velocities for the lower-body controller 30ms ahead.

\subsection{XBG Training}
XBG Training was conducted on a workstation equipped with two NVIDIA A100 PCIe 80GB (VRAM 160Gb), 1081GB of RAM, and an AMD EPYC 7513 32-Core Processor (128 parallel threads). The neural network was implemented in PyTorch, utilizing CUDA for GPU acceleration. Training and online deployment scripts were managed through Python and can be found in the paper repository: https://github.com/ami-iit/paper\_cardenas\_2024\_ral\_xbg.

Training uses the following configuration: Batch Size of 64, Learning Rate (LR) 0.0003, Adam optimizer and Mean Squared Error (MSE) as the Loss Function. 

We observed that the network with the lowest training or validation loss does not necessarily lead to the best behaviour performance. Consequently, the network's performance was evaluated based on the success ratio in HRI scenarios. We save and test the models during different epochs of the training, results presented correspond to the best of them with a maximum of 20 epochs which represent approximately 5 days of training.


\subsection{Real Robot Test}
To evaluate the system's performance, we conducted real-world tests on the ergoCub robot. The system operates at 30Hz, although internally, the network uses data sampled at 10Hz to align with the training setup. It's important to note that the controllers operate at a much higher frequency (500Hz). Additionally, walking velocities with lower value than 20\% of the maximum walking speed are consider as zero to avoid small steps with no locomotion.

An interaction is considered unsuccessful if the robot does not respond at all to the human interaction, responds incorrectly with a not accurate behaviour, or responds with a proper behaviour but fails to fully complete the interaction (i.e. if the robot let the payload fall during the walking with payload scenario, it is considered as a failure even if it reacts and walks in the right direction). Its worth to mention that the performance evaluation involved individuals who were not present in the training dataset. 
We trained different models conducting an ablation study on the effect of different inputs and components in the system performance. This study compares the model variations presented in table \ref{ablation} which presents the success ratio over 10 consecutive trial for each action. 

\begin{table}[ht]
\begin{center}
\begin{tabular}{|c||c|c|c|c|c|}
\hline
\textbf{Extero-} & \textbf{Hand-} & \textbf{Hand-} & \textbf{Payload} & \textbf{Walk} & \textbf{Walk with} \\ 
\textbf{ceptive} & \textbf{wave} & \textbf{shake} & \textbf{Reception} & \textbf{} & \textbf{payload} \\ \hline
RGB & 100\% & 30\% & 100\% & 80\% & 40\%\\ \hline
Depth & 0\% & 0\% & 0\% & 0\% & 0\%\\ \hline
\textbf{RGBD} & \textbf{100\%} & \textbf{60\%} & \textbf{90\%} & \textbf{100\%} & \textbf{100\%}\\ \hline
\end{tabular}
\end{center}
\caption{Ablation study: success over 10 trials}
\label{ablation}
\end{table}

As can be observed in table \ref{ablation}, the RGBD architecture presents a higher performance reason why we carried out further tests with 30 attemps for each behaviour, the success of each one are: handwave (100\%), handshake (53\%), payload reception (100\%), walking (93\%) and walking with payload (90\%).

\subsection{Behaviour blending}

In this experiment, we aimed to evaluate the model's ability to transition seamlessly between different actions randomly, simulating real-world scenarios where the robot may need to perform various tasks consecutively without explicit instruction.
We randomly selected a sequence of 30 interactions from our set of scenarios and executed them one after another. The robot autonomously determined how to behave based on its understanding of the environment and the changing context, successfully completing 70\% of these scenarios (See the associated video). 

This experiment provides valuable insights into the model's generalization capabilities and its ability to adapt to dynamic and unpredictable real-world scenarios. It highlights the system's robustness and flexibility in diverse human-robot interaction contexts, demonstrating the potential of our approach in handling complex, varied tasks seamlessly.

\section{CONCLUSIONS}\label{CONCLUSIONS}

In this study, we have developed and evaluated an end-to-end autonomous system for humanoid robots, aimed at real-world Human-Robot Interaction scenarios. Through teleoperation and careful data collection, we have constructed a comprehensive dataset encompassing diverse HRI interactions. Our approach integrates exteroceptive and proprioceptive components, providing the robot with a significant understanding of its environment and actions. We have addressed challenges such as data synchronization, downsampling, and noise reduction to optimize system performance. Real-world tests on the ergoCub robot demonstrate the effectiveness of our system, with actions repeated either consecutive or random to ensure robust evaluation. By focusing on correct behaviour completion, we provide a thorough assessment of the system's capabilities. 
Possible future work includes integrating new modalities to the XBG model such as the audio perceived by the robot, exploiting the potential of the Large Language Models (LLMs) in Human-Robot Interactions.

\addtolength{\textheight}{-12cm}   





\section*{ACKNOWLEDGMENT}

Thanks to all Artificial and Mechanical Intelligence members at IIT who helped us during the different experiments conducted.


\bibliographystyle{IEEEtran}

\bibliography{main}

\begin{thebibliography}{10}
\providecommand{\url}[1]{#1}
\csname url@samestyle\endcsname
\providecommand{\newblock}{\relax}
\providecommand{\bibinfo}[2]{#2}
\providecommand{\BIBentrySTDinterwordspacing}{\spaceskip=0pt\relax}
\providecommand{\BIBentryALTinterwordstretchfactor}{4}
\providecommand{\BIBentryALTinterwordspacing}{\spaceskip=\fontdimen2\font plus
\BIBentryALTinterwordstretchfactor\fontdimen3\font minus \fontdimen4\font\relax}
\providecommand{\BIBforeignlanguage}[2]{{%
\expandafter\ifx\csname l@#1\endcsname\relax
\typeout{** WARNING: IEEEtran.bst: No hyphenation pattern has been}%
\typeout{** loaded for the language `#1'. Using the pattern for}%
\typeout{** the default language instead.}%
\else
\language=\csname l@#1\endcsname
\fi
#2}}
\providecommand{\BIBdecl}{\relax}
\BIBdecl

\bibitem{Darvish2023teleoperation}
K.~Darvish, L.~Penco, J.~Ramos, R.~Cisneros, J.~Pratt, E.~Yoshida, S.~Ivaldi, and D.~Pucci, ``Teleoperation of humanoid robots: A survey,'' \emph{IEEE Transactions on Robotics}, vol.~PP, 06 2023.

\bibitem{SEMERARO2023102432}
\BIBentryALTinterwordspacing
F.~Semeraro, A.~Griffiths, and A.~Cangelosi, ``Human–robot collaboration and machine learning: A systematic review of recent research,'' \emph{Robotics and Computer-Integrated Manufacturing}, vol.~79, p. 102432, 2023. [Online]. Available: \url{https://www.sciencedirect.com/science/article/pii/S0736584522001156}
\BIBentrySTDinterwordspacing

\bibitem{abaigbena2024aihri}
\BIBentryALTinterwordspacing
A.~Obaigbena, O.~A. Lottu, E.~D. Ugwuanyi, B.~S. Jacks, E.~O. Sodiya, and O.~D. Daraojimba, ``Ai and human-robot interaction: A review of recent advances and challenges,'' \emph{GSC Advanced Research and Reviews}, 2024. [Online]. Available: \url{https://gsconlinepress.com/journals/gscarr/content/ai-and-human-robot-interaction-review-recent-advances-and-challenges}
\BIBentrySTDinterwordspacing

\bibitem{rozo2016controllers}
L.~Rozo, J.~Silvério, S.~Calinon, and D.~Caldwell, ``Learning controllers for reactive and proactive behaviors in human-robot collaboration,'' \emph{Frontiers in Robotics and AI}, vol.~3, pp. 1--11, 06 2016.

\bibitem{luo2023104414}
\BIBentryALTinterwordspacing
J.~Luo, W.~Liu, W.~Qi, J.~Hu, J.~Chen, and C.~Yang, ``A vision-based virtual fixture with robot learning for teleoperation,'' \emph{Robotics and Autonomous Systems}, vol. 164, p. 104414, 2023. [Online]. Available: \url{https://www.sciencedirect.com/science/article/pii/S0921889023000532}
\BIBentrySTDinterwordspacing

\bibitem{ogata2022cognitive}
\BIBentryALTinterwordspacing
T.~Ogata, K.~Takahashi, T.~Yamada, S.~Murata, and K.~Sasaki, ``{Machine Learning for Cognitive Robotics},'' in \emph{{Cognitive Robotics}}.\hskip 1em plus 0.5em minus 0.4em\relax The MIT Press, 05 2022. [Online]. Available: \url{https://doi.org/10.7551/mitpress/13780.003.0014}
\BIBentrySTDinterwordspacing

\bibitem{Zhang2017DeepIL}
\BIBentryALTinterwordspacing
T.~Zhang, Z.~McCarthy, O.~Jow, D.~Lee, K.~Goldberg, and P.~Abbeel, ``Deep imitation learning for complex manipulation tasks from virtual reality teleoperation,'' in \emph{IEEE International Conference on Robotics and Automation}, 2017. [Online]. Available: \url{https://api.semanticscholar.org/CorpusID:3720790}
\BIBentrySTDinterwordspacing

\bibitem{codevilla2022mutimodal}
\BIBentryALTinterwordspacing
Y.~Xiao, F.~Codevilla, A.~Gurram, O.~Urfalioglu, and A.~M. L\'{o}pez, ``Multimodal end-to-end autonomous driving,'' \emph{IEEE Transactions on Intelligent Transportation Systems}, vol.~23, no.~1, p. 537–547, jan 2022. [Online]. Available: \url{https://doi.org/10.1109/TITS.2020.3013234}
\BIBentrySTDinterwordspacing

\bibitem{perez2022visually}
\BIBentryALTinterwordspacing
R.~Pérez-Dattari, B.~Brito, O.~{de Groot}, J.~Kober, and J.~Alonso-Mora, ``Visually-guided motion planning for autonomous driving from interactive demonstrations,'' \emph{Engineering Applications of Artificial Intelligence}, vol. 116, p. 105277, 2022. [Online]. Available: \url{https://www.sciencedirect.com/science/article/pii/S0952197622003323}
\BIBentrySTDinterwordspacing

\bibitem{sergey2018vision}
\BIBentryALTinterwordspacing
R.~Rahmatizadeh, P.~Abolghasemi, L.~B\"{o}l\"{o}ni, and S.~Levine, ``Vision-based multi-task manipulation for inexpensive robots using end-to-end learning from demonstration,'' in \emph{2018 IEEE International Conference on Robotics and Automation (ICRA)}.\hskip 1em plus 0.5em minus 0.4em\relax IEEE Press, 2018, p. 3758–3765. [Online]. Available: \url{https://doi.org/10.1109/ICRA.2018.8461076}
\BIBentrySTDinterwordspacing

\bibitem{Mandlekar2020git}
\BIBentryALTinterwordspacing
A.~Mandlekar, D.~Xu, R.~Mart{\'{\i}}n{-}Mart{\'{\i}}n, S.~Savarese, and L.~Fei{-}Fei, ``Learning to generalize across long-horizon tasks from human demonstrations,'' \emph{CoRR}, vol. abs/2003.06085, 2020. [Online]. Available: \url{https://arxiv.org/abs/2003.06085}
\BIBentrySTDinterwordspacing

\bibitem{nasiriany2022learning}
S.~Nasiriany, T.~Gao, A.~Mandlekar, and Y.~Zhu, ``Learning and retrieval from prior data for skill-based imitation learning,'' 2022.

\bibitem{finn2017oneshot}
C.~Finn, T.~Yu, T.~Zhang, P.~Abbeel, and S.~Levine, ``One-shot visual imitation learning via meta-learning,'' 2017.

\bibitem{mandlekar2021whatmatters}
\BIBentryALTinterwordspacing
A.~Mandlekar, D.~Xu, J.~Wong, S.~Nasiriany, C.~Wang, R.~Kulkarni, L.~Fei-Fei, S.~Savarese, Y.~Zhu, and R.~Mart{\'\i}n-Mart{\'\i}n, ``What matters in learning from offline human demonstrations for robot manipulation,'' in \emph{5th Annual Conference on Robot Learning}, 2021. [Online]. Available: \url{https://openreview.net/forum?id=JrsfBJtDFdI}
\BIBentrySTDinterwordspacing

\bibitem{Dosovitskiy2020AnII}
\BIBentryALTinterwordspacing
A.~Dosovitskiy, L.~Beyer, A.~Kolesnikov, D.~Weissenborn, X.~Zhai, T.~Unterthiner, M.~Dehghani, M.~Minderer, G.~Heigold, S.~Gelly, J.~Uszkoreit, and N.~Houlsby, ``An image is worth 16x16 words: Transformers for image recognition at scale,'' \emph{ArXiv}, vol. abs/2010.11929, 2020. [Online]. Available: \url{https://api.semanticscholar.org/CorpusID:225039882}
\BIBentrySTDinterwordspacing

\bibitem{zhu2022viola}
Y.~Zhu, A.~Joshi, P.~Stone, and Y.~Zhu, ``Viola: Imitation learning for vision-based manipulation with object proposal priors,'' \emph{6th Annual Conference on Robot Learning (CoRL)}, 2022.

\bibitem{seo2023trill}
M.~Seo, S.~Han, K.~Sim, S.~H. Bang, C.~Gonzalez, L.~Sentis, and Y.~Zhu, ``Deep imitation learning for humanoid loco-manipulation through human teleoperation,'' in \emph{IEEE-RAS International Conference on Humanoid Robots (Humanoids)}, 2023.

\bibitem{robinson2023hric}
\BIBentryALTinterwordspacing
N.~Robinson, B.~Tidd, D.~Campbell, D.~Kuli\'{c}, and P.~Corke, ``Robotic vision for human-robot interaction and collaboration: A survey and systematic review,'' \emph{J. Hum.-Robot Interact.}, vol.~12, no.~1, feb 2023. [Online]. Available: \url{https://doi.org/10.1145/3570731}
\BIBentrySTDinterwordspacing

\bibitem{dafarra2024icub3}
\BIBentryALTinterwordspacing
S.~Dafarra, U.~Pattacini, G.~Romualdi, L.~Rapetti, R.~Grieco, K.~Darvish, G.~Milani, E.~Valli, I.~Sorrentino, P.~M. Viceconte, A.~Scalzo, S.~Traversaro, C.~Sartore, M.~Elobaid, N.~Guedelha, C.~Herron, A.~Leonessa, F.~Draicchio, G.~Metta, M.~Maggiali, and D.~Pucci, ``icub3 avatar system: Enabling remote fully immersive embodiment of humanoid robots,'' \emph{Science Robotics}, vol.~9, no.~86, p. eadh3834, 2024. [Online]. Available: \url{https://www.science.org/doi/abs/10.1126/scirobotics.adh3834}
\BIBentrySTDinterwordspacing

\bibitem{iFeel}
\BIBentryALTinterwordspacing
``ifeel.'' [Online]. Available: \url{https://ifeeltech.eu/}
\BIBentrySTDinterwordspacing

\bibitem{kourosh2019retargeting}
K.~Darvish, Y.~Tirupachuri, G.~Romualdi, L.~Rapetti, D.~Ferigo, F.~J.~A. Chavez, and D.~Pucci, ``Whole-body geometric retargeting for humanoid robots,'' in \emph{2019 IEEE-RAS 19th International Conference on Humanoid Robots (Humanoids)}, 2019, pp. 679--686.

\bibitem{romualdi2020benchmarking}
G.~Romualdi, S.~Dafarra, Y.~Hu, P.~Ramadoss, F.~J.~A. Chavez, S.~Traversaro, and D.~Pucci, ``A benchmarking of dcm-based architectures for position, velocity and torque-controlled humanoid robots,'' \emph{International Journal of Humanoid Robotics}, vol.~17, no.~01, p. 1950034, 2020.

\bibitem{yolov8_ultralytics}
\BIBentryALTinterwordspacing
G.~Jocher, A.~Chaurasia, and J.~Qiu, ``Ultralytics yolov8,'' 2023. [Online]. Available: \url{https://github.com/ultralytics/ultralytics}
\BIBentrySTDinterwordspacing

\bibitem{ergoCub}
\BIBentryALTinterwordspacing
``ergocub.'' [Online]. Available: \url{https://ergocub.eu/}
\BIBentrySTDinterwordspacing

\end{thebibliography}

\end{document}